\documentclass[pdflatex,sn-mathphys-num,iicol]{sn-jnl}

\usepackage{graphicx}%
\usepackage{multirow}%
\usepackage{amsmath,amssymb,amsfonts}%
\usepackage{amsthm}%
\usepackage{mathrsfs}%
\usepackage[title]{appendix}%
\usepackage{xcolor}%
\usepackage{textcomp}%
\usepackage{manyfoot}%
\usepackage{booktabs}%
\usepackage{algorithm}%
\usepackage{algorithmicx}%
\usepackage{algpseudocode}%
\usepackage{listings}%
\usepackage[normalem]{ulem}

\usepackage{subcaption}
\usepackage{tabularx}
\usepackage{xcolor}
\usepackage{lmodern}

\definecolor{lightgreen}{RGB}{54, 123, 47}
\definecolor{lightblue}{RGB}{57, 57, 166}
\definecolor{lightred}{RGB}{173, 80, 75}

\newcommand{\modify}[1]{{#1}}

\newcommand{\eg}[0]{\textit{e.g.}}   

\newcolumntype{C}{>{\centering\arraybackslash}X}

\begin{document}

\title[Article Title]{Rethinking Functional Brain Connectome Analysis: \mbox{Do Graph Deep Learning Models Help}}

\author[1]{\fnm{Keqi} \sur{Han}}\email{keqi.han@emory.edu}

\author[2]{\fnm{Yao} \sur{Su}}\email{ysu6@wpi.edu}

\author[3]{\fnm{Lifang} \sur{He}}\email{lih319@lehigh.edu}

\author[4]{\fnm{Liang} \sur{Zhan}}\email{liang.zhan@pitt.edu}

\author[5]{\fnm{Sergey} \sur{Plis}}\email{splis@gsu.edu}

\author[5]{\fnm{Vince} \sur{Calhoun}}\email{vcalhoun@gsu.edu}

\author*[1]{\fnm{Carl} \sur{Yang}}\email{j.carlyang@emory.edu}

\affil*[1]{\orgdiv{Dept. of Computer Science}, \orgname{Emory University}, \orgaddress{\city{Atlanta}, \state{GA}, \country{USA}}}

\affil[2]{\orgdiv{Dept. of Computer Science}, \orgname{ Worcester Polytechnic Institute}, \orgaddress{\city{Worcester}, \state{MA}, \country{USA}}}

\affil[3]{\orgdiv{Dept. of Computer Science and Engineering}, \orgname{Lehigh University}, \orgaddress{\city{Bethlehem}, \state{PA}, \country{USA}}}

\affil[4]{\orgdiv{Dept. of Electrical and Computer Engineering}, \orgname{Univ. of Pittsburgh}, \orgaddress{\city{Pittsburgh}, \state{PA}, \country{USA}}}

\affil[5]{\orgdiv{TReNDS Center}, \orgname{GSU, Georgia Tech, Emory University}, \orgaddress{\city{Atlanta}, \state{GA}, \country{USA}}}

\abstract{Graph deep learning models, a class of AI-driven approaches employing a message aggregation mechanism, have gained popularity for analyzing the functional brain connectome in neuroimaging. However, their actual effectiveness remains unclear. In this study, we re-examine graph deep learning versus classical machine learning models based on four large-scale neuroimaging studies. Surprisingly, we find that the message aggregation mechanism, a hallmark of graph deep learning models, does not help with predictive performance as typically assumed, but rather consistently degrades it. 
To address this issue, we propose a hybrid model combining a linear model with a graph attention network through dual pathways, achieving robust predictions and enhanced interpretability by revealing both localized and global neural connectivity patterns. Our findings urge caution in adopting complex deep learning models for functional brain connectome analysis, emphasizing the need for rigorous experimental designs to establish tangible performance gains and perhaps more importantly, to pursue improvements in model interpretability.}


\maketitle

\section{Introduction}\label{sec1}
Brain connectome analysis has captivated researchers for its potential to unravel brain organization and predict clinical outcomes \cite{wang2019hierarchical, long_range1, braingnn, bnt}. Functional magnetic resonance imaging (fMRI) plays a crucial role in this field by capturing brain activity through blood-oxygen-level-dependent (BOLD) signals. These signals facilitate the construction of functional brain networks, where nodes correspond to specific brain regions of interest (ROIs) and edges represent pair-wise statistical relationships, typically correlations, between the BOLD signals of these regions \cite{smith2011network, simpson2013analyzing, edge_centric, complex_measures}. Studying the functional connectome can enhance our understanding of brain functions \cite{dmn_ce2, vim_sm2}, enabling us to identify neural circuitry patterns associated with neurological conditions \cite{vim_sm3, pos_neg2} and develop improved diagnostic and therapeutic strategies \cite{sm1, sym2, medial_temp2, mood_disorder}.

Machine learning (ML) models have long been employed in functional brain connectome analysis, showing strong robustness and interpretability \cite{pos_neg3, pos_neg2, finn2015functional, shen2017using, cui2018effect, PERVAIZ2020116604, cmep}. These models use the correlation strength between ROIs as predictive features, without considering the underlying graph structure. As AI increasingly permeates healthcare and neuroscience, graph deep learning (GDL) models like graph neural networks (GNNs) \cite{braingnn, braingb, ibgnn, fbnetgen, neurograph} and graph transformers (GTs) \cite{bnt, gated, community_bnt} have recently gained popularity in this field. They explicitly model brain networks as graphs, aggregating messages among nodes (ROIs) along edges (functional connections) \cite{mpnn}. Among various node features, the connection profile---rows of the functional connection matrix---is the most common choice \cite{braingb, fbnetgen, bnt, ibgnn, braingnn, neurograph, gated, community_bnt, hypergraph}. Despite significant efforts to develop GDL models for brain connectome analysis, there is unfortunately no rigorous study and clear understanding of the genuine effectiveness of these advanced models over classical ML models, particularly regarding their unique message aggregation mechanisms.

In this work, we conduct a comprehensive evaluation of GDL against classical ML models, across four fMRI-based brain connectome datasets: ABIDE \cite{abide}, PNC \cite{pnc}, HCP \cite{hcp}, and ABCD \cite{abcd}. These datasets cover a diverse population and clinical prediction tasks, from demographic attributes to cognitive abilities and neurological disorders. The experimental results are surprising: (1) classical ML models, such as logistic regression and kernel ridge regression, along with simple multi-layer perceptrons (MLPs), often match or even exceed the predictive performance of more complex GDL models; and (2) the message aggregation mechanisms unique to GDL models and believed to be a key technical advantage, appear to have a consistent detrimental effect on prediction performance within the context of functional brain networks. These results challenge prevailing assumptions about the superiority of GDL methods in modeling brain connectome.

Motivated by the effectiveness of classical models and the limitations of GDL, we introduce a simple yet effective model featuring a novel dual-pathway design, which synergizes linear modeling (LM) with a graph attention network (GAT) \cite{gat}. The LM pathway efficiently captures global connectivity patterns from whole connectivity matrices, while the GAT pathway adaptively extracts local graph structures using embeddings of BOLD time series as node features. Despite its simple architecture, experimental results demonstrate that our model consistently achieves robust and competitive performance across diverse datasets, rivaling or surpassing the best-performing baselines. Beyond predictive performance, the model offers more holistic and complementary interpretability through its dual-pathway design: the GAT pathway is particularly effective at highlighting localized subnetworks and functional hubs, offering insights into modularized brain organization, while the LM pathway emphasizes global connectivity patterns and network-wide efficiency. The patterns identified not only align well with established neuroscientific knowledge but also reveal specific functional connections, salient ROIs, and subnetworks associated with cognitive processes, potentially providing finer-grained insights into neural mechanisms.

The findings of this study suggest caution with aggregation-based GDL models, particularly when connection profiles are used as node features, where aggregation shows clear negative effects. They underscore the importance of thoroughly benchmarking new models in future brain connectome research. Meanwhile, we advocate for future research to prioritize interpretability and contextual knowledge for scientific or clinical applications over simply pursuing prediction accuracy.

\begin{figure*}[!htbp]
    \centering

    \includegraphics[width=\textwidth]{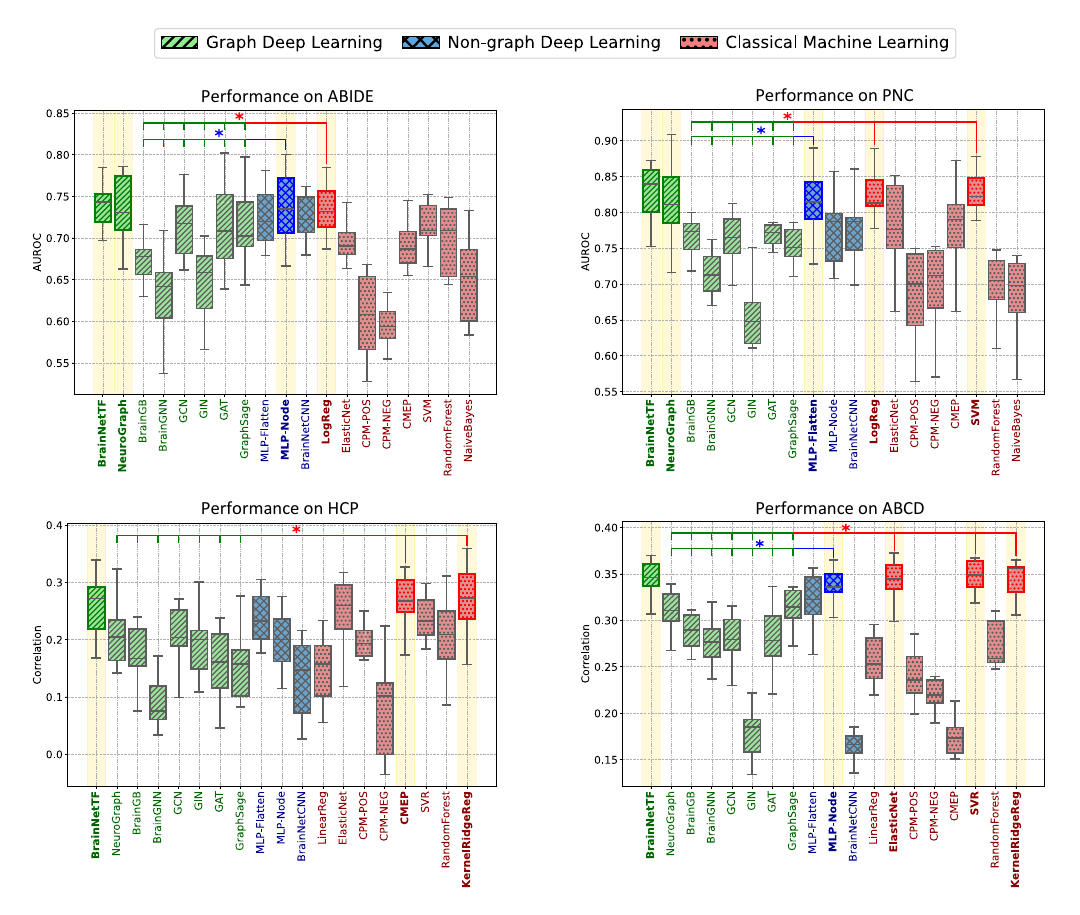}

    \caption{\textbf{Performance comparison across different baselines on four datasets.} For the ABIDE and PNC datasets, the prediction tasks are binary classification for autism disease and biological gender, respectively, evaluated using AUROC scores. For the HCP and ABCD datasets, the prediction task is fluid intelligence score prediction, a regression task evaluated using the Pearson correlation between predicted and actual values. The baselines are grouped into three categories, distinguished by color and hatching patterns: \textcolor{lightgreen}{green box for graph deep learning,} \textcolor{lightblue}{blue box for non‑graph deep learning,} and \textcolor{lightred}{red box for classical machine learning models}. Models highlighted with bold borders, colored hatching and yellow highlight band represent those with the best predictive performance among all the baselines. There is no significant difference ($p > 0.05$) in performance among these highlighted models, showing that simpler models such as Logistic Regression, MLP, ElasticNet, and Kernel Ridge Regression perform comparably to the most advanced graph deep learning models. The significance bars marked with asterisks ($*$) indicate that the performance of non-graph deep learning models is significantly better than that of the graph deep learning models displayed to the left ($p < 0.05$). The results show that simple models can match or even exceed the predictive performance of more complex GDL models.}
    \label{r1}
\end{figure*}

\section{Results}\label{sec2}

\noindent \textbf{Simple models can match or even exceed the predictive performance of more complex GDL models}

\vspace{0.5em}

To address the ambiguity regarding the effectiveness of complex GDL models, we conduct extensive experiments across four fMRI-based brain network datasets with various clinical prediction tasks, ranging from demographics to cognitive ability and neural disorders: ABIDE \cite{abide} for autism disease classification, PNC \cite{pnc} for gender prediction, and both HCP \cite{hcp} and ABCD \cite{abcd} for fluid intelligence prediction. Our benchmark spans three types of models: GDL, non-graph deep learning, and classical ML models.

Fig. \ref{r1} illustrates the prediction performance of each baseline across various datasets. Simpler models often match or outperform more advanced GDL models. Specifically, for classification tasks on ABIDE and PNC datasets, classical models (\eg, Logistic Regression) and MLP-based models consistently outperform most GDL models, with performance comparable to state-of-the-art models like NeuroGraph and BrainNetTF. Similarly, for regression tasks on HCP and ABCD datasets, classical ML models such as ElasticNet \cite{PERVAIZ2020116604} and Kernel Ridge Regression \cite{pos_neg2} demonstrate superior performance over most GDL models, and perform comparably to BrainNetTF \cite{bnt}, one of the most advanced aggregation-based GDL models. These findings challenge the prevailing belief that graph-based modeling of brain connectome is superior regarding the prediction of clinical outcomes. 

\vspace{1.8em}

\begin{figure*}[!htbp]
    \centering  
    
    \includegraphics[width=\textwidth]{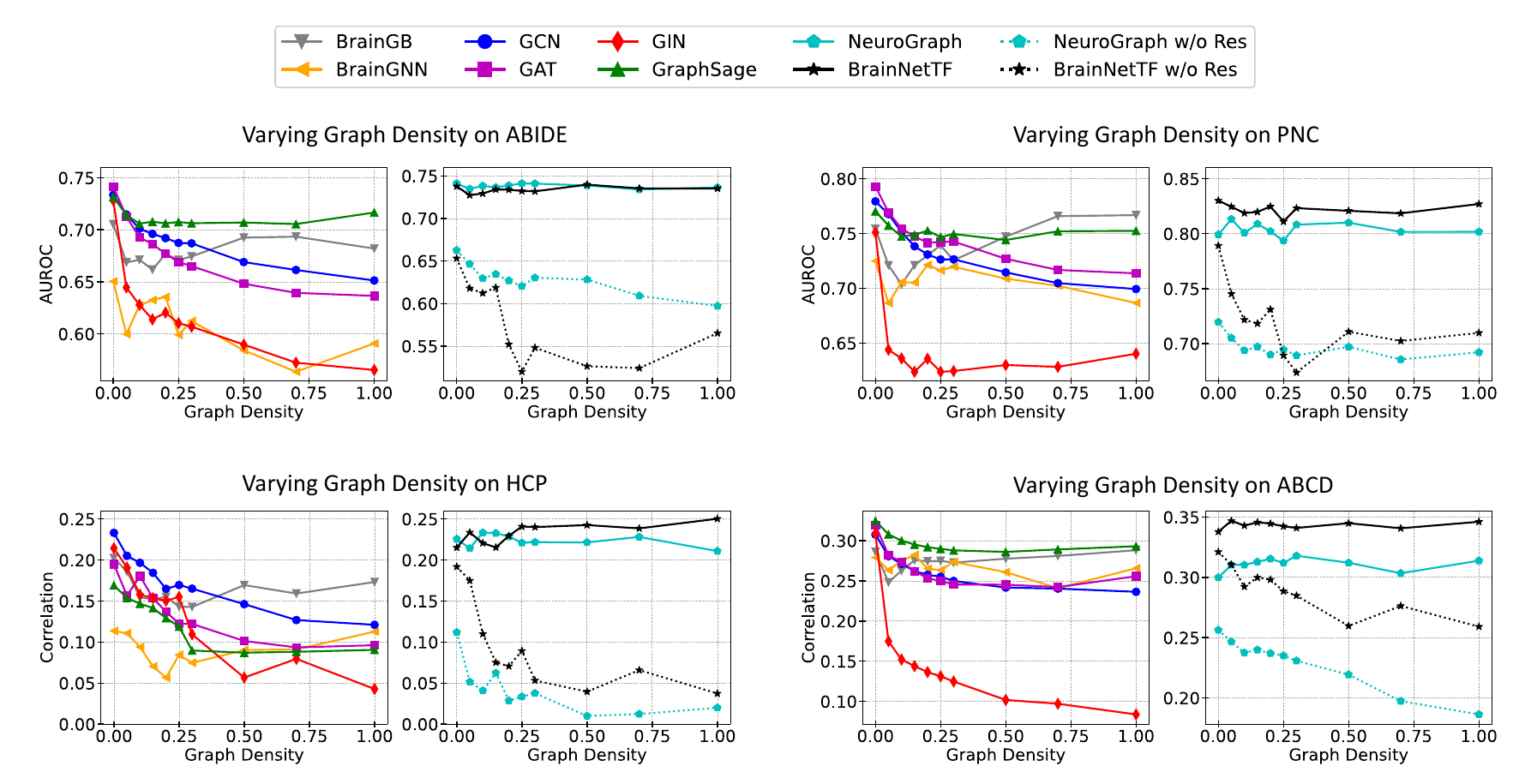}

    \vspace{0.5em}
    
    \caption{\textbf{Performance of graph deep learning models with varying graph densities.} To investigate the effect of aggregation in graph deep learning models on brain connectome analysis, we adjust graph density by retaining the top $K\%$ edges, thereby regulating the extent of aggregation between ROIs. Lower densities imply reduced aggregation. Specifically, when $K = 0$, no edges exist between ROIs, meaning node features are transformed independently without any aggregation. Each data point here represents the average result from 10 independent runs with the corresponding graph density. The left panel of each subfigure illustrates prediction accuracy trends of GCN, GAT, GIN, GraphSage, BrainGB, and BrainGNN, showing a consistent decline in performance as graph density increases. This indicates that message aggregation does not improve prediction outcomes but rather degrades performance. In contrast, the right panel of each subfigure highlights the performance of BrainNetTF and NeuroGraph, which remain stable across different graph densities due to the inclusion of residual connections. When residual connections are removed in their respective variants (BrainNetTF w/o Res and NeuroGraph w/o Res), performance declines similarly to models on the left, restating that aggregation alone harms prediction accuracy in brain connectome analysis.}
    \label{r2}
\end{figure*}

\noindent \textbf{Message aggregation in GDL models has a detrimental effect on prediction performance within the context of functional brain connectome analysis}

\vspace{0.5em}

In GDL models, the message aggregation is crucial for combining node features from neighboring nodes to form new node representations. 
However, as shown in Fig. \ref{r1}, most aggregation-based GDL models tend to underperform classical ML models like Logistic Regression, SVM/SVR \cite{hastie2009elements} and Kernel Ridge Regression, and simpler MLP-based models. This raises the question about the genuine effectiveness of the aggregation operations in brain connectome analysis. 

To further investigate the impact of aggregation, we vary the graph density by retaining different percentages of the top $K\%$ edges in the graphs, so as to explicitly control the extent of aggregation. Notably, with $K=0$, the graphs have no edges between ROIs, so node features are transformed independently, eliminating any aggregation effects. To align with existing studies, the original settings of each GDL model are followed for extracting the top $K\%$ edges. Specifically, for models like GCN \cite{gcn}, GAT \cite{gat}, GIN \cite{gin}, GraphSage \cite{graphsgae}, NeuroGraph \cite{neurograph}, and BrainGNN \cite{braingnn}, only positive correlations between ROIs are considered as potential edges. This approach adheres to the common assumption that positive correlations are typically more informative in the context of functional brain networks \cite{braingnn, neurograph}. Conversely, for BrainGB \cite{braingb} and BrainNetTF \cite{bnt}, both positive and negative correlations are incorporated, following the methodologies described in the original papers. 


The left panels of the Fig. \ref{r2} show that models like GCN, GAT, GIN, and GraphSage consistently exhibit a decline in prediction performance as the graph density increases, indicating that message aggregation does not enhance prediction and more aggregation can lead to more performance degradation. Similarly, BrainGB and BrainGNN demonstrate optimal performance with lower density, aligning with the notion that message aggregation may not be helpful in brain connectome analysis tasks.


Unlike other GDL models, BrainNetTF and NeuroGraph incorporate not only message aggregation but also residual connections, which integrate the untransformed input brain connectivity with the transformed hidden embeddings. The right panels of the Fig. \ref{r2} indicate that the performance of NeuroGraph and BrainNetTF remains relatively stable across varying graph density. However, when residuals are removed, as in the variants NeuroGraph w/o Res and BrainNetTF w/o Res, performance declines as density increases, consistent with other GDL models. This further underscores the point that message aggregation alone impairs predictive performance in brain connectome analysis.

\begin{figure}[t]
    \centering
    \includegraphics[width=\columnwidth]{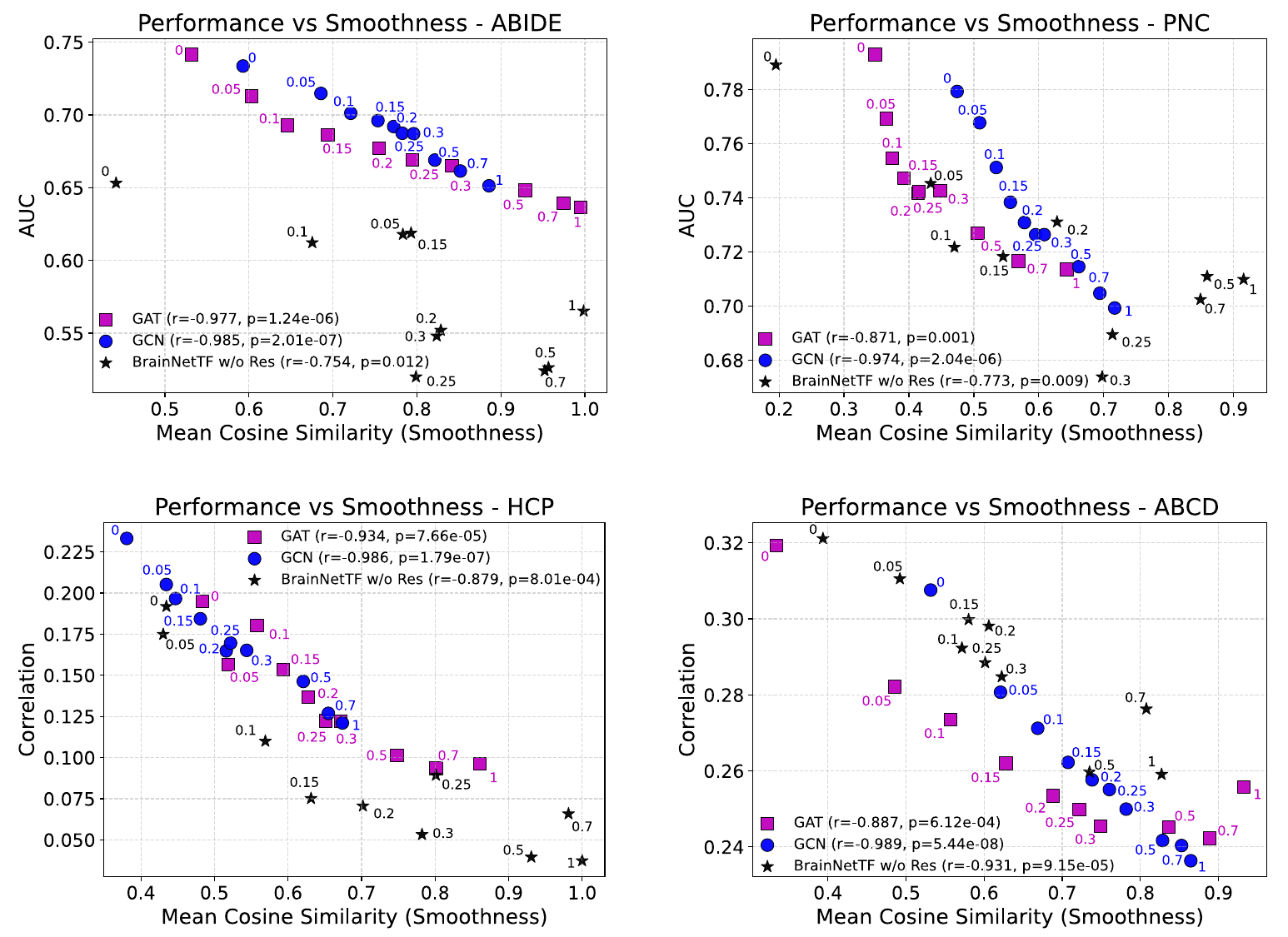}

    \vspace{10pt}
    \caption{\modify{\textbf{Relationship between node representation smoothness and predictive performance.} Each panel shows the relationship between model performance (y-axis) and the mean cosine similarity of the last-layer node embeddings (x-axis), which quantifies the smoothness of node representations. Results are presented for four datasets and three representative GDL models (GCN, GAT, and BrainNetTF without residual connections). Each point corresponds to the average over 10 independent runs under a specific graph density, with the adjacent number indicating the corresponding density level. Models are distinguished by color and marker type. The Pearson correlation coefficient between predictive performance and cosine similarity and its $p$-value are reported in the legend. Across all datasets and models, node smoothness and predictive performance exhibit a strong and statistically significant negative correlation (Pearson $r<0$, all $p<0.05$). As graph density increases, message aggregation leads to progressively smoother (more homogeneous) node representations, which coincide with decreased predictive accuracy. This indicates a manifestation of the over-smoothing phenomenon that underlies the limited effectiveness of GDLs in functional brain connectome analysis.}}
    \label{scatter_perf_sim}

\end{figure}

\modify{To further probe the mechanism behind the observed degradation in predictive performance, we conduct an additional analysis focusing on the smoothness of the learned node features by GDL models. Specifically, we examine three representative GDL models---GCN, GAT, and BrainNetTF (w/o residuals), which together capture the main architectural paradigms of graph convolutional networks and graph transformers. The residual connections in BrainNetTF are removed in this analysis to better isolate the effect of message aggregation itself. For each model and dataset, we compute the mean cosine similarity of the last-layer node embeddings (i.e., the representations directly used for prediction) across varying graph densities. This metric is widely adopted to measure the smoothness of representation \cite{chen2020measuring}, where higher values indicate that node embeddings become more homogeneous. As shown in Fig.~\ref{scatter_perf_sim}, across all four datasets and three representative GDL models, we consistently observe that as graph density increases, message aggregation generally leads to progressively smoother node features, accompanied by a decline in predictive accuracy. Moreover, the analysis reveals a strong and statistically significant negative correlation between node smoothness and model performance (Pearson $r < 0$ and $p < 0.05$). This consistent inverse relationship provides quantitative evidence that, in the context of functional brain connectome analysis, message aggregation tends to make node features converge and lose discriminability, a manifestation of the over-smoothing phenomenon that underlies the limited effectiveness of GDL models.}

\begin{table}[h]
\centering

\caption{\modify{\textbf{Effective rank of input node-feature matrices.}
For each dataset, the table reports the mean effective rank of the input node-feature matrices (averaged across subjects) together with the total feature dimensionality. The Rank/Dim ratio quantifies the proportion of independent variation within the input features. Across all datasets, the effective ranks are notably low (7–38\% of the feature dimensionality), indicating that the connection-profile-based node features lie in a low-dimensional subspace and contain substantial redundancy and correlation among nodes. This low-rank property contributes to the tendency of graph message aggregation to induce over-smoothing in learned representations.
}}

\begin{tabularx}{\columnwidth}{lCCC}
\toprule
\textbf{Dataset} & \textbf{Node Feature Dim} & \textbf{Effective Rank} & \textbf{Rank/Dim Ratio} \\
\midrule
ABIDE & 200 & 26.02 & 0.13 \\
PNC   & 264 & 19.35 & 0.07 \\
HCP   & 132 & 49.94 & 0.38 \\
ABCD  & 360 & 33.84 & 0.09 \\
\bottomrule
\end{tabularx}
\label{effective_rank}
\end{table}

\modify{In functional brain connectome analysis, GDL models typically use the functional connection profile as the node feature, which has become a standard practice. Each node feature corresponds to a row of the functional connectivity matrix and thus inherently encodes the to all other nodes in the network. Consequently, every node feature already carries global information, leading to substantial correlation and redundancy among nodes even before any message aggregation. To quantify this redundancy, we examine the effective rank \cite{roth2024rank} of the input node feature matrices for each dataset. As shown in Table~\ref{effective_rank}, the average effective ranks are notably low relative to the input dimensionality (7–38\% of feature dimensions). Given such low-rank and highly correlated inputs, message aggregation---which essentially acts as a low-pass filter over the graph \cite{li2018deeper}, further amplifies shared components among nodes and dilutes distinctive variations, making the learned representations more homogeneous and thus more prone to over-smoothing.}

\modify{Collectively, these findings indicate that in functional connectome analysis, the inherent low-rank property of brain connectivity features makes GDLs vulnerable to over-smoothing under message aggregation, which in turn degrades their predictive capability. This also explains why architectures with residual or skip connections (e.g., NeuroGraph \cite{neurograph} and BrainNetTF \cite{bnt}) remain more stable performance, as they preserve variance from the original, unaggregated node features and thereby mitigating over-smoothing.}

\vspace{1em}

\noindent \textbf{The proposed dual-pathway model achieves robust prediction performance across various datasets}

\vspace{0.5em}

Previous experiments have shown that simple linear models like Logistic Regression and ElasticNet, often outperform other baselines. However, relying solely on linear models overlooks graph structure information inherent in brain networks, which is essential for understanding brain function and cognitive processes \cite{complex_measures, clustering_coef1}. To incorporate this information, it is natural to consider using graph models. However, Fig. \ref{r2} reveals that using connection profiles as node features diminishes the effectiveness of aggregation operations within graph models.

To address these challenges, we propose a dual-pathway model that integrates the strengths of both linear non-graph and aggregation-based graph models. The model consists of two complementary pathways. The first pathway retains the simplicity and effectiveness of linear models, which uses the flattened upper triangular part of the functional brain network matrix as input. The second pathway utilizes graph structure by embedding BOLD time series, rather than connection profiles, as node features. \modify{These BOLD embeddings, which are then processed by a graph attention network (GAT), are obtained through a 1D-CNN encoder that captures local temporal dynamics of each ROI, providing more diverse and less redundant inputs for graph aggregation.}
 Table \ref{r3_performance} shows that this dual-pathway model outperforms baselines on the ABCD and PNC datasets, matches performance on ABIDE, and exhibits slight underperformance on HCP. Overall, this dual-pathway model achieves robust and competitive performance across four datasets.


\begin{table}[!h]
    \centering
    \caption{\textbf{Comparison of the dual-pathway model with the best-performing baseline across four datasets (mean $\pm$ std).}}
    
    \begin{tabularx}{\columnwidth}{l>{\centering\arraybackslash}X>{\centering\arraybackslash}X}
        \toprule
        \textbf{Dataset} & \textbf{Dual-pathway Model} & \textbf{Best-performing Baseline} \\
        \midrule
        ABIDE & 0.732 $\pm$ 0.037 & \textbf{0.737 $\pm$ 0.044} \\
        PNC & \textbf{0.834 $\pm$ 0.039} & 0.827 $\pm$ 0.042 \\
        HCP & 0.247 $\pm$ 0.062 & \textbf{0.270 $\pm$ 0.063} \\
        ABCD & \textbf{0.358 $\pm$ 0.019} & 0.348 $\pm$ 0.017 \\
        \bottomrule
    \end{tabularx}
    \label{r3_performance}
\end{table}

\vspace{1em}
\noindent \textbf{Two pathways of the proposed model reveal different and complementary neural circuitry patterns}

\vspace{0.5em}

The proposed dual-pathway model integrates a non-aggregation linear model (LM) and an aggregation-based graph attention network (GAT). Each pathway uncovers unique neural circuitry patterns linked to brain functions. The LM pathway offers interpretability through the weights assigned to the input features---vectorized upper triangular portion of the brain network matrix, while the GAT pathway provides interpretability via attention maps averaged across test samples.

\modify{In this section, we use the ABCD dataset as a representative case study to demonstrate the interpretability of the dual-pathway model. ABCD is selected for its large sample size (over 7,000 participants), which supports statistical stability and robust estimation of model-derived circuitry patterns \cite{marek2022reproducible}. It also employs a high-resolution 360-region parcellation, providing broad spatial coverage of ROIs and sufficient granularity to identify interpretable neural systems.} To help understand the circuitry patterns captured by two pathways, we map ROIs into six commonly recognized neural systems based on structural and functional roles \cite{functional_module}: the Somatomotor Network (SM), Default Mode Network (DMN), Ventral Salience Network (VS), Central Executive Network (CE), Dorsal Salience Network (DS), and Visual Network (Vis).

We systematically examine the revealed neural circuitry from three aspects: \textit{Important Connections (Edge-level)}, \textit{Salient ROIs (Node-level)}, and \textit{Graph Properties of the Highlighted Brain Connectivity (Subgraph-level)}. Detailed discussions on these discoveries are deferred to Section 3.

\vspace{1em}

\begin{figure*}[!htbp]
    \centering

    \includegraphics[width=\textwidth]{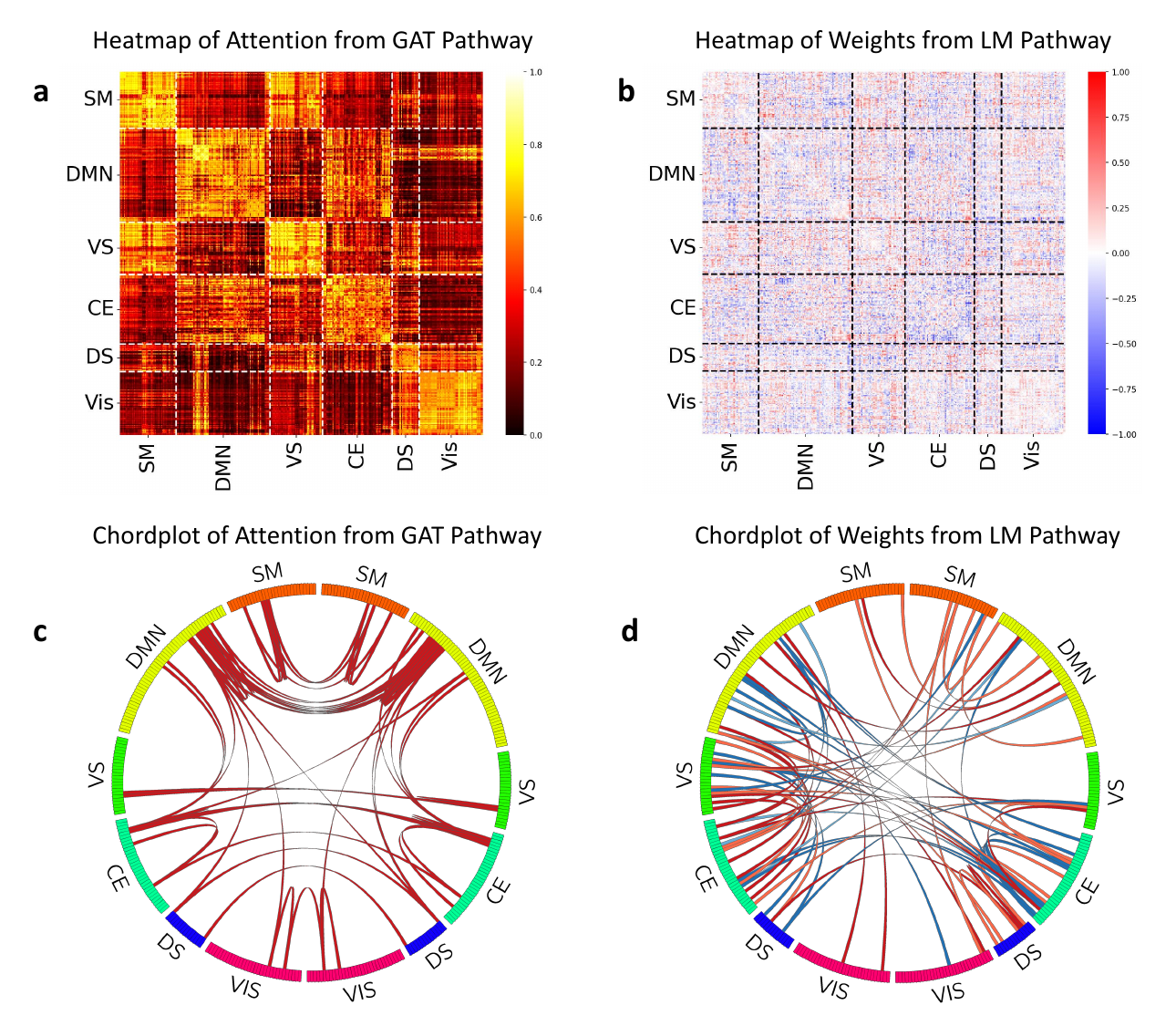}

    \vspace{1 em}
    
    \caption{\textbf{Important connections identified by two pathways in the ABCD dataset.} \textbf{(a)-(b)} Heatmap of important connections for GAT and LM pathways. The attention scores from the GAT pathway are rescaled to [0,1] using min-max normalization, while weights from the LM pathway are scaled to [-1,1] by normalizing with the maximum absolute value of weights. The heatmaps show that the GAT pathway highlights within-system connections, particularly in the Default Mode (DMN), Somatomotor (SM) and Central Executive (CE) networks, aligning with their roles in cognitive processes. In contrast, the LM pathway reveals more uniformly distributed connections, which reflects the long-range network integration essential for cognitive flexibility and large-scale communication. \textbf{(c)-(d)} Chordplots of the top $0.1\%$ most significant connections (65 in total) for GAT and LM pathways. The GAT pathway reveals symmetrical connections between homologous ROIs across hemispheres, highlighting its focus on both local and bilateral brain organization. In contrast, the LM pathway shows a more diffusive pattern with both positive and negative weights distributed across networks, accounting for both facilitative and inhibitory interactions across ROIs, necessary for maintaining cognitive flexibility.}
    
    \label{r3_important_connections}
\end{figure*}

\noindent Important Connections (Edge-level): To explore the relationship between functional brain networks and cognitive abilities, we begin by visualizing the interpretability results through heatmaps. Fig. \ref{r3_important_connections}(a) presents the attention scores from the GAT pathway. It addresses prominent within-system connections, indicated by the diagonal blocks. The cross-system interactions between DMN-CE and Vis-SM also stand out. In contrast, Fig. \ref{r3_important_connections}(b) displays the weights from the LM pathway. This heatmap includes both positive (red) and negative (blue) weights, reflecting the direction and magnitude of each connection's impact on the prediction. Compared to the attention map, the weights are more uniformly distributed, indicating that the LM pathway captures broader connections beyond local interactions.

\vspace{0.5em}

To complement the global views from the heatmaps, we focus on the top $0.1\%$ of the most significant connections (65 in total). Fig. \ref{r3_important_connections}(c) shows the chordplot from the GAT pathway, which emphasizes within-system connectivity, especially within the DMN, SM, and CE systems. There are fewer cross-module connections, most of which involve the DMN. 
Additionally, there is a noticeable symmetry in the chordplot, with many connections linking the same ROI across the left and right hemispheres, underscoring GAT’s focus on both local and bilateral brain network organizations. 
In contrast, Fig. \ref{r3_important_connections}(d) presents the chordplot from the LM pathway, where connections are more evenly spread across both within-system and cross-system interactions, with a balanced distribution of positive and negative weights.

 \begin{figure*}[t]
    \centering

    \includegraphics[width=\textwidth]{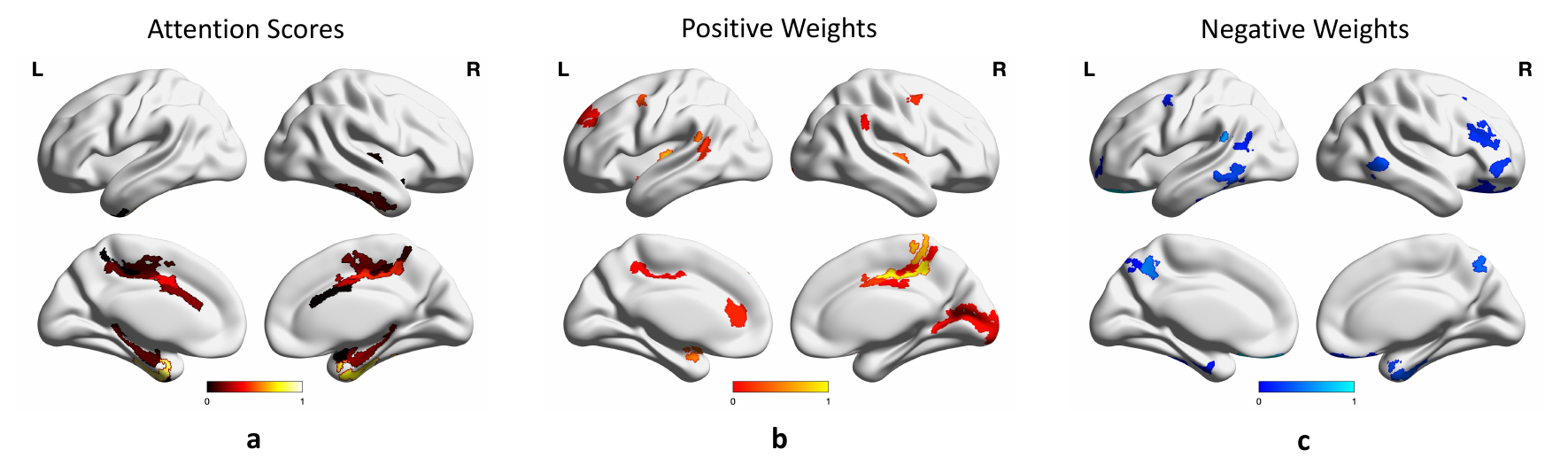}

    \vspace{0.5 em}
    
    \caption{\textbf{Visualization of the top 20 salient ROIs identified by two pathways in the ABCD dataset.} Color intensity represents the importance of each ROI, with darker shades indicating lower importance and brighter shades indicating higher importance. \textbf{(a)} For the GAT pathway, node importance is calculated by summing the attention scores across all its connections, corresponding to each row in the attention heatmap. The identified salient ROIs are primarily concentrated in the Central Executive Network (CE) and Default Mode Network (DMN), with key regions located in the frontal and temporal lobe. \textbf{(b)-(c)} For the LM pathway, node importance is calculated separately for positive and negative weights. Salient ROIs identified from the LM pathway’s positive weights are concentrated in the Ventral Salience Network (VS) and Somatomotor Network (SM), predominantly in the temporal lobe, while negative weights highlight regions within the DMN, particularly in the frontal and parietal lobes.}
    \label{r3_rois}
    \end{figure*}


\vspace{0.5em}

\noindent Salient ROIs (Node-level): For the GAT pathway, node importance is determined by summing the attention scores across all connections per node, while for the LM pathway, positive and negative weights are considered separately. We select the top 20 ROIs based on node importance and visualize their locations using the BrainNet Viewer tool \cite{brainnetviewer} in Fig. \ref{r3_rois}.


For the GAT pathway, the top 20 ROIs are concentrated within the CE and DMN system, with key regions in the frontal lobe, including the Orbital and Polar Frontal cortexes, and the Inferior Frontal cortex, as well as the Medial Temporal cortex in the temporal lobe.

For the LM pathway, the top 20 ROIs with positive weights are primarily concentrated in the VS and SM system, mainly located in the temporal lobe, particularly within the Insular and Frontal Opercular cortexes. Additionally, significant regions are also found in the parietal lobe, within the Paracentral Lobular and Mid-Cingulate cortexes. 
Negative weights primarily indicate ROIs in the DMN system, especially in the frontal and parietal lobes, such as the Orbital and Polar Frontal cortexes, Temporo-Parieto-Occipital Junction, and Posterior Cingulate cortexes.


\begin{figure*}
    \centering

    \includegraphics[width=\textwidth]{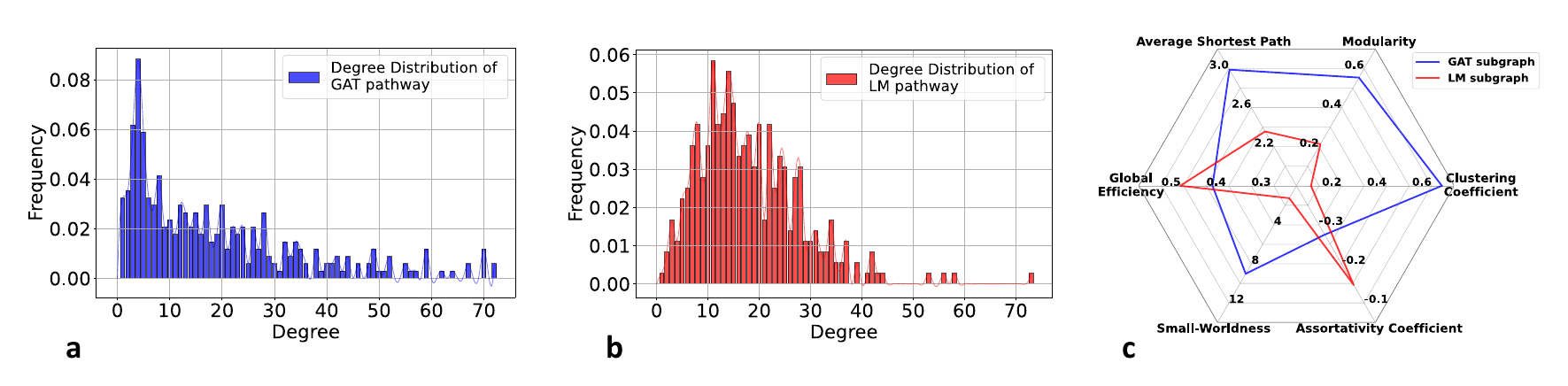}

    \vspace{0.5 em}

    \caption{\textbf{Graph properties of highlighted subgraphs derived from the top 5\% of edges from two pathways in the ABCD dataset.} \textbf{(a)-(b)} Node degree distribution of subgraphs for GAT and LM pathways. The GAT-induced subgraph exhibits a skewed distribution with a few highly connected hubs, indicative of a scale-free network, often associated with the brain's robustness and resilience to localized failures. The LM-induced subgraph exhibits a broader distribution with a higher concentration of nodes having intermediate degrees, reflecting the distributed processing across multiple brain regions. \textbf{(c)} Comparison between other numerical graph properties. The GAT-induced subgraph demonstrates higher clustering coefficients and modularity, indicating tightly-knit communities and clear modular divisions. Additionally, the small-worldness metric is significantly higher, suggesting an optimal balance between local specialization and global integration. In contrast, the LM-induced subgraph features shorter average path lengths and higher global efficiency, reflecting a network structure optimized for rapid information transfer. Besides, the higher assortativity coefficient in the LM subgraph points to stronger connections between functionally similar brain regions.}
    \label{r3_subgraph}
    \end{figure*}

\vspace{0.5em}

\noindent Graph Properties of the Highlighted Brain Connectivity (Subgraph-level): We now shift our attention to subgraph-level structural characteristics of the brain connectivities identified as important by our model. Specifically, subgraphs are constructed using the top 5\% of edges identified as important from the GAT and LM pathways respectively, and key graph properties \cite{complex_measures}---degree distribution, clustering coefficient, modularity, average shortest path length, global efficiency, small-worldness, and assortativity---are computed to compare the brain connectivities emphasized by each pathway.

The degree distribution, as shown in Fig. \ref{r3_subgraph}(a) and \ref{r3_subgraph}(b), reveals distinct patterns between the GAT-induced and LM-induced subgraphs. The GAT subgraph displays a more skewed distribution, indicating the presence of a few highly connected hubs, which are characteristics of scale-free networks. While the LM subgraph exhibits a broader distribution with a higher concentration of nodes having intermediate degrees, suggesting a normal distribution of connectivity.

Other graph properties are shown in Fig. \ref{r3_subgraph}(c). the GAT subgraph exhibits higher clustering coefficients and modularity, pointing to a graph structure that favors tightly-knit communities and clear modular divisions. The LM subgraph shows shorter average path lengths and higher global efficiency, indicative of an efficient network. Besides, the assortativity coefficient is higher in the LM subgraph, reflecting a stronger tendency for connections between similar nodes. Notably, the small-worldness metric, which reflects a balance between local clustering and global integration, is significantly higher in the GAT subgraph.

\vspace{1em}

\noindent The dual-pathway model demonstrates distinct yet complementary neural circuitry patterns across both pathways. The GAT pathway highlights prominent within-system connectivity and reveals a small-world structure with a strong local clustering phenomenon. The LM pathway shows more evenly distributed connections and a network structure emphasizing global efficiency and cross-system interactions. Together, these pathways offer a more holistic view of brain organization. The combined insights reinforce the model's potential to capture diverse neural mechanisms underlying cognitive and behavioral outcomes.

\section{Discussion}\label{sec3}


Despite advancements in GDL and the growing interest in aggregation-based models for functional brain connectome analysis \cite{braingb, fbnetgen, bnt, ibgnn, dag, braingnn, neurograph, gated, community_bnt}, our findings suggest that sophisticated GDL models do not necessarily outperform simpler models, such as classical ML models and MLPs. 

Many existing works implicitly assume that GDL's complexity guarantees superior results. In practice, these studies often omit direct comparisons with simpler models \cite{braingb, fbnetgen, bnt, community_bnt}, thereby leaving the true effectiveness of these complex models inadequately assessed. Even when simpler baselines are included, as in BrainGNN \cite{braingnn} and NeuroGraph \cite{neurograph}, benchmarking is incomplete, and their experimental conclusions are inconsistent. For instance, BrainGNN shows SVM and Random Forest outperforming graph models in some cases, while MLPs underperform; conversely, NeuroGraph finds MLPs significantly outperforming most graph models, with weaker Random Forest results. Our study address the limitations from these works by rigorously optimizing and benchmarking both simple and complex models, providing a more comprehensive evaluation of various models for brain connectome analysis.

Our results suggest that simple models, when properly tuned, can serve as strong baselines and, in some cases, even outperform more complex models. This highlights the need for a more balanced perspective where simple models are not only included as baselines but also carefully evaluated to ensure that any proposed advancements genuinely offer superior performance.

\vspace{0.5em}

Although message aggregation techniques are widely adopted across domains to capture complex relational information \cite{social_recommentation, he2020lightgcn, molecular}, in functional brain networks, our results show that these operations degrade rather than improve prediction accuracy.
GDL-based brain connectome analysis models typically use the connection profile as the node feature, which represents one row of the functional connectivity matrix and inherently encodes the correlations between a given region of interest (ROI) and all others \cite{braingb, fbnetgen, bnt, ibgnn, braingnn, neurograph, gated, community_bnt, hypergraph}. \modify{While standard in the field, this design implies that each node feature already captures global network information and shows considerable redundancy and inter-node correlation even before any aggregation.}

\modify{This intrinsic redundancy is quantified in Table.~\ref{effective_rank}, where the effective rank of the input node-feature matrices is found to be substantially lower than the feature dimensionality (7–38\% of feature dimensions across datasets), indicating that the input features reside in a low-dimensional subspace. In such a low-rank regime, message aggregation, which essentially acts as a low-pass filter \cite{li2018deeper} over the graph, tends to further amplify the shared components of node features and suppress distinctive variations, thereby progressively diminishing the predictive capability of the learned node representations, a process consistent with the over-smoothing \cite{smoothing} phenomenon observed in GDL models.}

\modify{These observations suggest a mismatch between the prevailing formulation of functional brain connectome analysis (Pearson correlation–based connectivity with connection profile node features) and the inductive bias of message aggregation in current GDL models. Improving the alignment between brain representation and model architecture may help enhance the effectiveness of GDLs applied to functional brain connectome analysis. Future studies could explore alternative graph formulations and model variants that better account for the redundancy and low-rank characteristics of functional connectivity data, aiming to capture more meaningful and predictive neural patterns.}


\vspace{0.5em}

Motivated by these observations, we develop a dual-pathway hybrid model that jointly leverages linear and graph-based representations to capture complementary aspects of brain connectivity. This hybrid model achieves robust predictive performance across datasets. The LM pathway leverages the strong predictive power inherent in the original functional brain connectivity, preserving the direct input-output relationship and avoiding feature distortions. Meanwhile, the GAT pathway improves upon previous models by using BOLD time series embeddings as node features, instead of connection profiles. BOLD time series embeddings retain the raw temporal dynamics and variability of ROIs' signals, which captures richer temporal information and mitigates over-smoothing issues linked to the strong dependencies and redundancy in connection profiles.


\vspace{0.5em}
To further understand how the dual-pathway model captures functional brain organization, we examined its interpretability from three complementary perspectives: edge-, node-, and subgraph-level patterns.

At the edge level, the GAT pathway’s focus on within-system connections, particularly within DMN, SM and CE, resonates with well-established findings in neuroscience about cognitive ability. The DMN is tied to internally directed cognitive processes such as self-referential thinking, memory retrieval, and future planning \cite{self_ref1, self_ref3, mem_ret3}. Similarly, the SM network, crucial for sensory processing, has shown within-network coherence related to cognitive abilities \cite{sm1, sm2}. The CE network’s within-system emphasis reinforces its critical role in executive functions like working memory, decision-making, and cognitive control, which is correlated with fluid intelligence \cite{ce2, ce4}. This within-system focus reflects the specialized, localized processing within functional networks. Moreover, the observed bilateral symmetry suggests that the GAT pathway effectively captures the functional coherence between homologous regions across hemispheres, which is known to be important for cognitive stability and performance \cite{sym2, sym3}. While the identification of cross-system connections between DMN-CE \cite{dmn_ce1, dmn_ce2}, as well as between the Vis-SM systems \cite{vim_sm2, vim_sm3}, underscores the importance of network integration in cognitive function. The LM pathway, on the other hand, captures a more diffusive and global connectivity pattern, balancing positive and negative weights across brain networks. This balance indicates LM pathway’s ability to account for both facilitative and inhibitory interactions across ROIs, necessary for maintaining cognitive flexibility \cite{pos_neg2, pos_neg3}. The dispersed weighting also reflects the integration of long-range connections necessary for large-scale network communication and distributed processing across the brain functional systems \cite{long_range1, long_range2, long_range3}. 

At the node level, the model highlights salient regions of interest (ROIs) that align well with established neural substrates of cognitive control and introspective processing. The concentration of attention-based ROIs within the CE and DMN aligns well with their well-documented roles in cognitive control and introspective processes. The involvement of the Orbital, Polar, and Inferior Frontal cortexes aligns with their contributions to executive functions and decision-making \cite{orbital_frontal1, orbital_frontal3, orbital_frontal4}. The identification of key regions within the Medial Temporal cortex by the GAT pathway also aligns with its known role in memory formation and retrieval \cite{medial_temp1, medial_temp2}, critical for reasoning and knowledge application. The engagement of both frontal and temporal lobes highlights the distributed nature of cognitive processes \cite{dist2} underlying fluid intelligence. The LM pathway adds further interpretability through its positive and negative weights. Positive weights, indicating regions where increased activity correlates with higher fluid intelligence, are prominent within the VS and SM systems. The VS, particularly involving the Insular and Frontal Opercular cortexes, is crucial for detecting and prioritizing relevant stimuli \cite{insular_frontal}, while the SM supports motor planning and cognitive execution \cite{sm1}. In contrast, negative weights, indicating that increased activity is associated with lower fluid intelligence, are found mainly in DMN, especially within the Orbital and Polar Frontal cortexes. This aligns with the ``Default Mode Interference Hypothesis'' \cite{hypothesis},  positing that excessive DMN activity, associated with mind-wandering and self-referential thought, can interfere with task-focused cognition.

At the subgraph level, the GAT-induced subgraph, the skewed degree distribution reflects the scale-free nature typical of biological networks. Such networks are known for their resilience to random failures, which supports the brain’s robustness in maintaining cognitive functions \cite{long_range1, scale_free1}. The higher clustering coefficient and modularity suggest that the attention mechanism captures tightly-knit communities, consistent with the brain's modular organization, where distinct networks support specialized functions \cite{complex_measures, clustering_coef1}. The notably higher small-worldness metric further highlights a balance between local specialization and global integration \cite{small_world1, small_world3}, a hallmark of efficient brain organization. 
Conversely, the LM-induced subgraph, with shorter average path lengths and higher global efficiency, suggests a network optimized for rapid information transfer and efficient brain communication \cite{communication2, communication3}. The broader degree distribution indicates a more even network \cite{dist2} without dominant hubs, possibly reflecting distributed processing across brain regions. Higher assortativity coefficient suggests a preference for connections between similar nodes \cite{complex_measures}, consistent with local homogeneity in certain brain areas \cite{homogeneity1, homogeneity2}.

In summary, the dual-pathway model offers a comprehensive view of brain connectivity, with each pathway contributing distinct interpretative information. The GAT pathway excels at capturing localized, modular structures, emphasizing key functional hubs and the brain's small-world properties. In contrast, the LM pathway highlights global integration and network efficiency, focusing on distributed processing and long-range connections. Together, the proposed hybrid model provides a balanced view, bridging specialized modular functions with broader network communication, thus deepening our understanding of the neural basis of brain functions. 
\modify{It is important to note that the interpretability analysis on the ABCD dataset is presented as a representative case study illustrating the dual-pathway model's ability to capture complementary local and global neural organization patterns, rather than to derive age-specific or cohort-dependent neuroscientific conclusions.}

\vspace{0.5em}

Finally, several limitations of this study should be acknowledged, along with potential avenues for future extensions.
First, We exclusively utilize rs-fMRI data to construct functional connectivity networks. However, the brain's structural connectivity, often derived from modalities like Diffusion Tensor Imaging (DTI), provides complementary information about the tangible physical pathways between ROIs. Integrating structural connectivity might offer a more holistic understanding of the role of GDL models for brain connectome analysis.

\modify{Second, in this study, we focus on the mainstream setting in which functional brain networks are constructed using Pearson correlation, following the most widely adopted practice in fMRI-based connectome analysis. All major observations and conclusions in this work are derived under this standard setting. 
Nonetheless, there still exist many other approaches to mapping functional connectivity \cite{liu2025benchmarking}, such as distance-based, more sophisticated causality-based, or spectral methods, as well as end-to-end adaptive graph learning strategies that infer connectivity during model training \cite{fbnetgen}, which we have not yet explored.
These alternative formulations might influence the effectiveness of GDL models in distinct ways and represent promising directions for future investigation.}

\modify{Third, this study primarily evaluates GDLs using two node-feature families: connection profiles and BOLD time-series embeddings. Results and conclusions are therefore grounded in these choices. Other options—e.g., graph-theoretical metrics, additional BOLD-derived handcrafted features, and ROI-specific neurological context—remain unexplored and may differentially affect GDL behavior and interpretability, motivating future studies in this direction.}

\section{Methods}\label{sec4}

\noindent \textbf{Dataset acquisition and processing}
\vspace{0.5em}


   The Autism Brain Imaging Data Exchange (ABIDE) dataset \cite{abide} is a publicly available dataset consists of anonymized resting-state functional magnetic resonance imaging (rs-fMRI) data collected from 17 international sites, providing a diverse and comprehensive collection for studying Autism Spectrum Disorder (ASD). The data is preprocessed using the C-PAC pipeline \cite{cpac}, which includes several stages: slice time correction using AFNI’s 3dTshift, motion correction with AFNI’s 3dvolreg, skull-stripping with AFNI’s 3dAutomask, global mean intensity normalization and band-pass filtering (0.01-0.1 Hz). After quality control, 1009 subjects are left, with 516 individuals (51.14\%) diagnosed with ASD. The prediction task focuses on binary classification for ASD diagnosis. The brain regions are defined using the Craddock 200 atlas \cite{Craddock_atlas}, with each subject represented by a $200 \times 200$ functional connectivity matrix. The BOLD time-series for each ROI are truncated to a length of 100.

    The Philadelphia Neuroimaging Cohort (PNC) \cite{pnc} originates from the Brain Behavior Laboratory at the University of Pennsylvania and the Children’s Hospital of Philadelphia. We preprocess this rs-fMRI dataset through a series of steps including slice timing correction, motion correction, registration, normalization, removal of linear trends, bandpass filtering (0.01-0.1 Hz), and spatial smoothing. After preprocessing and quality control, the dataset includes 504 subjects, with 289 (57.34\%) being female. The prediction task involves binary classification for gender prediction. The regions of interest are parcellated using the 264-node atlas defined by Power et al. \cite{power2011functional}. Each subject has time-series data collected through 120 time steps.

    The Human Connectome Project (HCP) \cite{hcp} is a systematic effort to map macroscopic human brain circuits and their relationship to behavior in a large population of healthy adults. For our analysis, we utilize resting-state fMRI data from the HCP dataset, processed using the standard pipeline in the CONN toolbox \cite{conn}, involving realignment, co-registration, normalization, and smoothing of the raw images. Confounding effects from motion artifacts, white matter, and cerebrospinal fluid (CSF) are regressed out of the signal. Subsequently, the BOLD time series are band-pass filtered (0.01-0.1 Hz) to retain relevant functional connectivity information. We focus on the 132 nodes defined on the Harvard-Oxford atlas \cite{desikan2006automated}. After rigorous quality control, this dataset includes 982 samples. Each sample is represented by a $132 \times 132$ functional connectivity matrix. The prediction task is fluid intelligence score prediction, treated as a regression problem, with the scores having a mean of 17.03 and a standard deviation of 4.70. To ensure consistency, the BOLD time-series are truncated to a length of 1024.

    The Adolescent Brain Cognitive Development Study (ABCD) \cite{abcd} is a large-scale, longitudinal project that recruits children aged 9-10 years across 21 sites in the United States. This study includes repeated imaging scans, along with extensive psychological and cognitive assessments. For our analysis, we use rs-fMRI data from the baseline visit, which is processed using the standard, open-source ABCD-HCP BIDS fMRI pipeline\footnote{https://github.com/DCAN-Labs/abcd-hcp-pipeline}. The preprocessing pipeline includes multiple stages, starting with distortion correction, anatomical alignment, and surface reconstruction. Further, the functional images undergo motion correction, distortion removal and registration. ANTs-based bias field correction and global signal regression are used to enhance data quality. The time series are band-pass filtered between 0.008 and 0.09 Hz. After quality control, we get 7,327 samples, each represented by a $360 \times 360$ connectivity matrix, constructed by calculating the Pearson correlation between the BOLD time-series of each pair of ROIs. The regions of interest (ROIs) are defined based on the HCP 360 atlas \cite{hcp360}. The prediction task is fluid intelligence score prediction with a mean of 18.21 and a standard deviation of 3.69, treated as a regression problem. Considering the variability in sequence lengths within the ABCD dataset, the BOLD time-series are truncated to a unified length of 512 for consistency.
    

\modify{We construct functional connectivity (FC) using Pearson correlation between regional fMRI time series, following the predominant practice in fMRI-based connectomics research \cite{braingb,bnt,ibgnn,neurograph,braingnn,pos_neg3,finn2015functional,cmep}.
This choice maintains methodological continuity with prior work and facilitates direct comparison and interpretation by the broader neuroimaging community.}

\modify{We intentionally select four fMRI datasets with distinct acquisition protocols, preprocessing pipelines, parcellation schemes, sample sizes and prediction tasks to evaluate the robustness and generalizability of our findings. All analyses are performed independently within each dataset, and conclusions are drawn based on consistent relative model performance patterns rather than absolute cross-dataset comparisons. This design ensures that the observed trends are not tied to a specific preprocessing pipeline or participant population, but instead reflect a more general property of GDL models in functional brain connectome analysis.}

\vspace{1.5em}
\noindent \textbf{Benchmarked baselines}
\vspace{0.5em}

For the classical machine learning baselines, we include Connectome-based Predictive Modeling (CPM) \cite{shen2017using}, a widely recognized model in network neuroscience for predicting behavioral or clinical outcomes based on brain connectivity data. CPM is typically split into two variations: CPM-NEG and CPM-POS, where predictive features are selected based on negative and positive correlations with the target variable, respectively. Covariance Maximizing Eigenvector-based Predictive Modeling (CMEP) \cite{cmep}, a more recent model that first generates candidate features based on eigenvectors with strong linear covariance with the target variable. These features are then used in ElasticNet regression to assess their predictive power, streamlining the process by avoiding arbitrary feature selection thresholds. Additionally, we employed classical models including Logistic Regression for classification, Linear Regression for regression, ElasticNet for both classification and regression, Support Vector Machine (SVM) for classification, Support Vector Regression (SVR) for regression, Random Forest for both classification and regression, as well as Naive Bayes for classification and Kernel Ridge Regression for regression. An overview of these models can be found in \cite{hastie2009elements}.

These models are chosen for their established use in brain connectome analysis, ensuring that our findings could be directly compared and validated against existing literature. For all models except CMEP, the input features are the vectorized upper triangular part of the functional brain network matrix, a setting commonly used in classical ML-based functional brain network analysis studies \cite{pos_neg3, pos_neg2, finn2015functional, shen2017using, cui2018effect, PERVAIZ2020116604}. CMEP uses the entire symmetric functional brain network matrix as input.

Most of these classical ML models require a feature selection process. For CPM, feature selection follows the original methodology by applying a p-value threshold \cite{shen2017using} to the correlation between each connection (feature) and the target variable. Features with p-values below the threshold are retained for model training and prediction. CMEP, however, does not require explicit feature selection as it inherently identifies predictive features during its eigenvector generation process. For the other classical ML models, feature selection involves selecting the top $m$ features with the smallest p-values. Specifically, for classification tasks (\eg, using the ABIDE and PNC datasets), p-values are determined through a t-test, while for regression tasks, they are calculated based on the correlation between features and the target variable.

\vspace{0.5em}
For the non-graph deep learning baselines, we adopt BrainNetCNN \cite{brainnetcnn}, a convolutional neural network (CNN) framework specifically designed for predicting clinical neurodevelopmental outcomes from brain networks. Unlike classical image-based CNNs that use spatially local convolutions, BrainNetCNN incorporates novel convolutional filters that operate on the topological locality of structural brain networks. These filters include edge-to-edge, edge-to-node, and node-to-graph convolutions, making the model particularly well-suited for leveraging the unique structure of brain connectivity data. Additionally, we also use two multi-layer perceptron (MLP)-based models: MLP-Flatten and MLP-Node. MLP-Flatten takes the upper triangular part of the brain connectome matrix as input, flattens it into a feature vector, and processes it through two fully connected layers. In contrast, MLP-Node transforms each node's feature individually using shared weights across nodes, then concatenates these transformed node features into a graph-level embedding for the final prediction.

\vspace{0.5em}

For the aggregation-based graph deep learning baselines, we focus on models that rely on message aggregation, a core mechanism through which nodes integrate information from their neighbors to update their representations. This process is critical for capturing both structural and relational properties in graph data. Our study includes general-purpose graph deep learning architectures as well as models specifically adapted to the characteristics of brain connectomes.

The general-purpose GDL baselines include following models: Graph Convolutional Network (GCN) \cite{gcn}, a fundamental graph neural network (GNN) model that generalizes convolutional operations to graph structures. It aggregates the features of neighboring nodes by applying a convolution-like operation, which computes a weighted sum of the neighboring node features. Graph Attention Network (GAT) \cite{gat} introduces an attention mechanism, allowing the model to assign different importance to neighboring nodes. The attention coefficients are adaptively learned during training, enabling the network to focus on the most relevant neighbors. This adaptive aggregation strategy makes GAT particularly useful for capturing the heterogeneity in node connections, where certain connections might be more relevant to the task. Graph Isomorphism Network (GIN) \cite{gin} is designed to be as powerful as the Weisfeiler-Lehman graph isomorphism test. It aggregates neighboring node features by applying a multi-layer perceptron (MLP) to the sum of the features. GraphSAGE \cite{graphsgae} updates node embeddings by averaging the features of neighboring nodes and passing the result through an MLP. 

In addition to these general models, specialized GDL models tailored for brain connectome analysis are also evaluated. BrainGB \cite{braingb} is a systematic study of how to design effective GNNs for brain connectome analysis. The core design dimensions of BrainGB include message passing mechanism and pooling strategies. BrainGNN \cite{braingnn} introduces ROI-aware graph convolutional layers that leverage both topological and functional information from fMRI data. The model also includes ROI-selection pooling layers (R-pool) to highlight the most salient ROIs for interpretability. Regularization terms are used to encourage biologically meaningful ROI selection. NeuroGraph \cite{neurograph} builds on a standard GCN by incorporating a residual connection. It concatenates the hidden embeddings from each GCN layer with the flattened upper triangular part of the brain network matrix. The resulting embedding undergoes batch normalization and is passed through an MLP for final prediction. BrainNetTF \cite{bnt} applies Transformer-based models to brain connectome analysis, treating each node in the brain network as a `word in a sentence'. This model utilizes the classic Transformer encoder, which captures pairwise connection strengths among ROIs through self-attention mechanism. Additionally, BrainNetTF introduces an Orthonormal Clustering Readout (OCRead) module, which performs self-supervised soft clustering and orthonormal projection to generate cluster-aware graph embeddings. 

\modify{All GDL baselines utilize connection profiles as node features, i.e., a row of the Pearson correlation–based functional connectivity matrix. This follows established practice in GDL studies \cite{neurograph, braingb, braingnn, bnt}, ensuring methodological alignment with prior work and enabling direct comparability and interpretation within the neuroimaging literature.}

For the input graphs of each GDL baseline, the original configurations from prior research are adhered to. Specifically, for models such as GCN, GAT, GIN, GraphSage, NeuroGraph, and BrainGNN, only positive correlations between ROIs are retained as edges, based on the common assumption that positive correlations are generally more informative in the context of functional brain networks \cite{braingnn, neurograph}. In contrast, for BrainGB and BrainNetTF, both positive and negative correlations are incorporated, as outlined in their original papers. The default graph densities for each GDL baseline also follow their original settings, with the default densities set to the top 5\% for GCN, GAT, GIN, GraphSage, and NeuroGraph, and the top 10\% for BrainGNN. For BrainGB and BrainNetTF, we follow their original settings and use the full graph with 100\% density.


Besides the baselines mentioned above, there are other methods related to our study but are not included as baselines. BrainNetGNN \cite{a14030075} and FBNetGen \cite{fbnetgen}, for instance, incorporate time series as part of node features but adaptively construct brain networks during learning. Since our study focuses on static networks, these methods are excluded. Additionally, recent transformer-based approaches, such as Community-aware BNT \cite{community_bnt} and Gated BNT \cite{gated}, require prior knowledge during training, such as neural system assignments or spatial coordinates for each ROI. As our study uses solely BOLD time-series data and the network matrices derived from them, we do not consider these methods as baselines.

\begin{figure}[ht]
    \centering
    \includegraphics[width=\columnwidth]{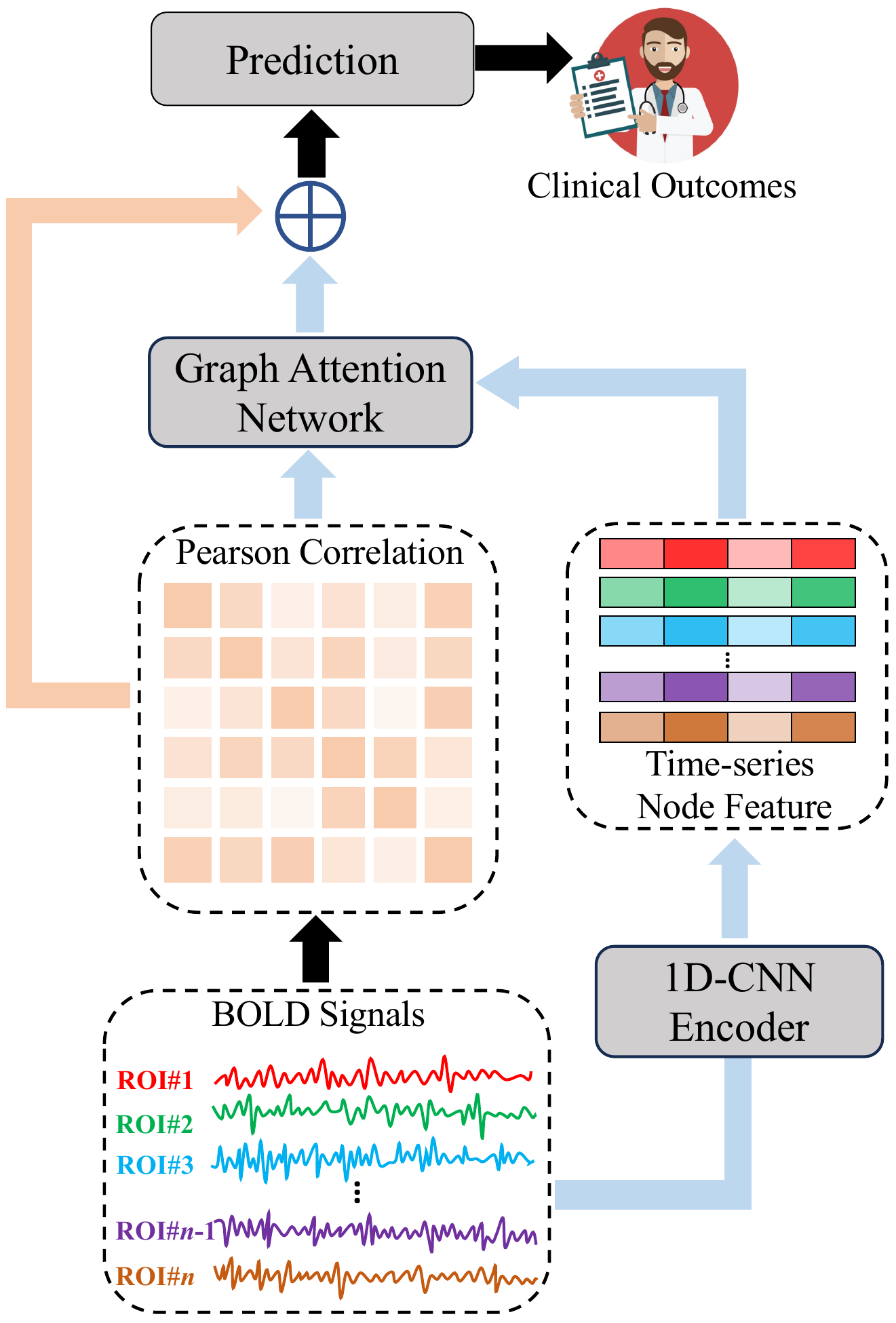}
    \caption{\textbf{Framework of the Dual-pathway model.} Orange arrow indicates LM pathway, while blue arrows represent GAT pathway.}
    \label{framework_dual}
\end{figure}

\noindent \textbf{Dual-pathway Model}
\vspace{0.5em}

The framework of the proposed dual-pathway model is depicted in Fig. \ref{framework_dual}, which comprises two complementary pathways. The first pathway (orange arrow) retains the simplicity and effectiveness of linear models (LM), using the flattened upper triangular part of the functional brain network matrix as its feature. Formally, for a given sample, let $\mathbf{B} \in \mathbb{R}^{n \times t}$ represent the BOLD time series data, where $n$ is the number of ROIs, and $t$ is the number of time points. The functional brain network is then represented by the Pearson correlation matrix $\mathbf{A} \in \mathbb{R}^{n \times n}$, where each element $\mathbf{A}_{ij}$ is the Pearson correlation coefficient between the time series of ROI $i$ and ROI $j$. The feature vector $\mathbf{u} \in \mathbb{R}^{\frac{n(n-1)}{2}}$ is obtained by flattening the upper triangular part of $\mathbf{A}$, which serves as the feature to the LM pathway. This feature construction approach is consistent with the settings in classical ML-based brain connectome analysis studies \cite{pos_neg3, pos_neg2, finn2015functional, shen2017using, cui2018effect, PERVAIZ2020116604}, which enables the model to capture global patterns from the whole connectivity matrices.

The second pathway (blue arrows) captures the structure information within brain connectome. It utilizes the functional brain network as the graph, \modify{with node features derived from the BOLD time series embeddings. This choice of node feature is motivated by our finding that connection profiles, which already encode global relationships among nodes, are highly redundant and thus prone to over-smoothing during message aggregation. In contrast, BOLD time-series embeddings preserve localized and complementary temporal dynamics of each ROI, enabling more effective information propagation across the graph. The localized information contained in the BOLD time-series embeddings also complement the global information modeled in the Linear Modeling (LM) pathway.} To obtain these embeddings, the BOLD time series data $\mathbf{B}$ is fed into a 1D-CNN encoder, resulting in node features $\mathbf{X} \in \mathbb{R}^{n \times d}$, where $d$ is the dimension of the time-series embeddings. \modify{We choose 1D-CNNs because they are computationally lightweight, less prone to overfitting given the typically limited sample sizes in neuroimaging datasets, and have been widely used in prior studies for modeling fMRI BOLD signals \cite {zhao2025advances}.}

The input graph $\hat{\mathbf{A}}$ for the GAT pathway is derived from the Pearson correlation matrix $\mathbf{A}$ by retaining only the positive correlations:

\begin{equation}
\hat{\mathbf{A}}_{ij} = 
\begin{cases}
\mathbf{A}_{ij}, & \text{if } \mathbf{A}_{ij} > 0, \\
0, & \text{otherwise}.
\end{cases}
\end{equation}



\noindent Following previous works \cite{braingnn, neurograph}, the negative connections are excluded for graph model. This step emphasizes the functional connectivity between ROIs that exhibit co-activation or co-deactivation, reflecting homophily, where the behavior patterns of connected ROIs are similar. This homophily aligns with the core assumptions of commonly used GDLs \cite{homophily}. In contrast, negative correlations between ROIs, which indicate heterophily or dissimilar behavior patterns, do not fit these assumptions.


The GAT model takes the graph $\hat{\mathbf{A}}$ and node features $\mathbf{X}$ as inputs to compute the node embeddings $\mathbf{H} \in \mathbb{R}^{n \times d'}$, where $d'$ is the dimension of the learned node embedding. A graph-level embedding $\mathbf{h}_{\text{graph}} \in \mathbb{R}^{d''},d''=n \times d'$ is then obtained via concatenation pooling: $\mathbf{h}_{\text{graph}} = \text{ConcatPooling}(\mathbf{H})$.


Finally, the graph-level embedding $\mathbf{h}_{\text{graph}}$ from the GAT pathway is concatenated with the vectorized feature vector $\mathbf{u}$ from the LM pathway: $\mathbf{z} = [\mathbf{h}_{\text{graph}}; \mathbf{u}]$, which is then passed through a fully connected layer to make the final prediction. Specifically, for classification tasks, the prediction is given by $\hat{y} = \sigma(\mathbf{W}\mathbf{z} + b)$, where $\sigma(\cdot)$ is softmax function. For regression tasks, the prediction is given by $\hat{y} = \mathbf{W}\mathbf{z} + b$, where $\mathbf{W}$ and $b$ represent the weights and bias of the fully connected layer.


Given the dominance of the LM pathway for prediction, there is a tendency for the GAT pathway to be undertrained during joint optimization. To mitigate this imbalance, we adopt a phased training strategy. Initially, the LM pathway is frozen, allowing the GAT pathway to train independently for a tunable number of epochs. Then, both pathways are jointly optimized, ensuring the GAT pathway is well-trained and contributes effectively to the overall prediction.

\vspace{0.5 em}
\noindent \textbf{Other Experimental Settings}
\vspace{0.2 em}

     For the classification tasks on the ABIDE and PNC datasets, model performance is evaluated using the AUROC score. For the regression tasks on the ABCD and HCP datasets, we use the Pearson correlation coefficient $r$ between the predicted scores and true scores as the performance metric.
     
     We randomly split each dataset into three subsets: 70\% for training, 10\% for validation, and 20\% for testing. All deep learning models are trained for 100 epochs using Adam optimizer with a batch size of 16 and a weight decay of 1e-4. The epoch with the highest AUROC (for classification tasks) or correlation (for regression tasks) on the validation set is used to compare performance on the test set. All reported performances are the average of 10 random runs on the test set with the standard deviation.

    We use the authors’ open-source codes for CPM, CMEP, BrainGB, BrainGNN, NeuroGraph, and BrainNetTF. The graph-based models GCN, GAT, GIN, and GraphSAGE are from the PyTorch Geometric library \cite{pyg}, while the classical ML baselines are the implementation from scikit-learn library \cite{scikit_learn}. Hyperparameter tuning is conducted via grid search for some important hyper-parameters of these baselines, based on the performance on the validation set. Specifically, for Dual-pathway model, we tune the number of independent training epochs for GAT pathway from \{10, 20, 50\}, number of layers \{2, 3, 4\}, and hidden channels \{32, 64, 128\}; for GCN, we tune the number of layers \{2, 3\}, hidden channels \{32, 128, 256\}, and the graph pooling method \{concatenation, mean pooling\}; for GAT, we adjust the number of layers \{2, 3\}, hidden channels \{32, 128, 256\}, and the number of attention heads \{2, 4\}; for GIN, we vary the number of layers \{2, 3\}, hidden channels \{32, 128, 256\}, and $\epsilon$ for neighborhood aggregation \{0.0, 0.2, 0.5\}; for GraphSAGE, we explore the number of layers \{2, 3\}, hidden channels \{32, 128, 256\}, and aggregation method \{mean, max\}; for BrainGB, we search for the best pooling strategy \{concatenation, mean pooling\}, hidden dimensions \{64, 128, 256\}, and message-passing type \{weighted sum, node concatenation\}; for BrainGNN, we fine-tune the feature dimensions \{32, 128\} and regularization coefficients \{0, 0.1\}; for NeuroGraph, the number of layers \{2, 3\} and hidden channels \{32, 64, 128\} are optimized; for BrainNetTF, the number of attention heads \{2, 4\} and clusters \{5, 10, 100\} are adjusted; for BrainNetCNN, we tune the first and second layer dimensions \{16, 32\} and \{32, 64\}, respectively, and the dropout rate \{0.3, 0.5\}; for all the deep learning models, we also tune the learning rate from \{1e-4, 5e-4, 1e-3, 1e-2\}; for CPM, the p-value threshold is varied \{0.001, 0.01, 0.05, 0.1, 0.5, 1\}; for CMEP, the elastic net regularization is fine-tuned with $\alpha$ values \{0.01, 0.1, 0.5\} and L1 ratios \{0.01, 0.05, 0.1, 0.5\}; for Logistic Regression, the regularization strength \{0.1, 1, 10\} and solver \{liblinear, lbfgs, saga\} are optimized; for ElasticNet, the $\alpha$ values \{0.001, 0.01, 0.1, 0.5, 1\} and L1 ratios \{0.2, 0.5, 0.7\} are adjusted; for SVM/SVR, we select between linear and RBF kernels; for Random Forest, the number of trees \{100, 200\} and maximum depth \{10, unlimited\} are fine-tuned; and for Kernel Ridge Regression, the $\alpha$ values \{0.1, 1, 10\} and kernel types \{linear, RBF, polynomial\} are explored. Additionally, for classical ML models other than CPM and CMEP, the number of the selected features ranges within \{100, 200, 500, 1000, 5000, 10000, n(n-1)/2\}, where $n$ is the number of ROIs.
    

\vspace{1em}
\noindent \textbf{Data availability} 
\vspace{0.2em}

\noindent The ABCD data is publicly available via the NIMH Data Archive (NDA) at \url{https://nda.nih.gov/abcd}, and the HCP data through ConnectomeDB at \url{https://db.humanconnectome.org/}. The PNC data can be accessed through the Philadelphia Neurodevelopmental Cohort (PNC) initiative at \url{https://www.nitrc.org/projects/pnc/}. Access to the ABCD, HCP, and PNC datasets requires an application. The ABIDE data is publicly accessible without restrictions through the Preprocessed Connectomes Project (PCP) at \url{http://preprocessed-connectomes-project.org/abide/}. This study does not involve any new datasets. All datasets used, along with their preprocessing steps, are properly cited within the text.

\vspace{1em}
\noindent \textbf{Code availability} 

\vspace{0.2em}

\noindent The code for the baseline experiments and the proposed dual-pathway model, along with detailed experimental settings and parameters used in each part of the study, can all be found at \url{https://github.com/LearningKeqi/RethinkingBCA}.

\vspace{0.8em}
\noindent \textbf{Author contributions}
\vspace{0.2em}

\noindent K.H. discovered key insights, conceived the project idea, designed the experiments, led the development of the dual-pathway model, and wrote the manuscript with input from all co-authors. C.Y. conceptualized and supervised the whole project, designed the methodology, provided resources (computing, funding, data), and guided manuscript writing. Y.S. and L.Z. participated in discussions, handled data preprocessing and contributed to writing. L.H. contributed to experimental design, discussions, and manuscript refinement. S.P. and V.C. provided suggestions on experimental analysis and manuscript writing, and contributed ideas that inspired the project.

\vspace{1em}
\noindent \textbf{Acknowledgements} 
\vspace{0.2em}

\noindent This research was supported in part by the US National Science Foundation under Award Numbers 2319449, 2312502, and 2442172, the internal funds and GPU servers provided by the Computer Science Department of Emory University, as well as the University Research Committee of Emory University.

\bibliography{05_References}

\end{document}